\pdfoutput=1

\documentclass[11pt]{article}

\usepackage[preprint]{acl}

\usepackage{times}
\usepackage{latexsym}
\usepackage{multirow}
\usepackage{booktabs}
\usepackage{bbding}
\usepackage[most]{tcolorbox}
\usepackage{CJKutf8}
\usepackage[T1]{fontenc}

\usepackage[utf8]{inputenc}

\usepackage{microtype}

\usepackage{inconsolata}

\usepackage{graphicx}
\usepackage{enumitem}
\title{MADial-Bench: Towards Real-world Evaluation of Memory-Augmented Dialogue Generation }

\author{Junqing He\textsuperscript{1}, Liang Zhu\textsuperscript{2}, Rui Wang\textsuperscript{1}, Xi Wang\textsuperscript{3}, Reza Haffari\textsuperscript{4}, Jiaxing Zhang\textsuperscript{1} \\
        \textsuperscript{1}International Digital Economy Academy, Shenzhen, China \\
        \textsuperscript{2}Southern University of Science and Technology, Shenzhen, China \\ 
        \textsuperscript{3}Univerity of Sheffield, Sheffield, U.K\\
        \textsuperscript{4}Monash University, Melbourne, Australia\\
        \texttt{\{hejunqing,wangrui\}@idea.edu.cn}}

\begin{document}
\maketitle
\begin{abstract}


Long-term memory is important for chatbots and dialogue systems (DS) to create consistent and human-like conversations, evidenced by numerous developed memory-augmented DS (MADS).
To evaluate the effectiveness of such MADS, 
existing commonly used evaluation metrics, like retrieval accuracy and perplexity (PPL), mainly focus on query-oriented factualness and language quality assessment. However, these metrics often lack practical value. Moreover, the evaluation dimensions are insufficient for human-like assessment in DS.
Regarding memory-recalling paradigms, current evaluation schemes only consider passive memory retrieval while ignoring diverse memory recall with rich triggering factors, e.g., emotions and surroundings, which can be essential in emotional support scenarios.
To bridge the gap, we construct a novel \textbf{M}emory-\textbf{A}ugmented \textbf{Dial}ogue \textbf{Bench}mark (\textbf{MADail-Bench}) covering various memory-recalling paradigms based on cognitive science and psychology theories. The benchmark assesses two tasks separately: \textit{memory retrieval} and \textit{memory recognition} with the incorporation of both passive and proactive memory recall data. We introduce new scoring criteria to the evaluation, including \textit{memory injection}, \textit{emotion support (ES) proficiency}, and \textit{intimacy}, to comprehensively assess generated responses. Results from cutting-edge embedding models and large language models on this benchmark indicate the potential for further advancement. Extensive testing further reveals correlations between memory injection, ES proficiency, and intimacy\footnote{Our code and dataset will be released upon acceptance.}.
\end{abstract}

\section{Introduction}
%
Long-term memory is crucial for achieving human-like communication in scenarios that demand sustained interactions, such as personal companionship, psychological counselling, and personal assistant tasks \cite{zhong2023memorybank}. Powerful Large Language Models (LLM) such as LlaMA \cite{touvron2023llama}, Claude \cite{claude3}, GPT4 \cite{achiam2023gpt} and ChatGPT \cite{ChatGPT} have demonstrated a remarkable ability to understand and generate coherent responses \cite{bubeck2023sparks}. However, their lack of long-term memory hinders their application in dialogue systems for natural conversations.


Therefore, researchers built diverse Memory-Augmented Dialogue Systems (MADS) that maintain contextual understanding, and ensure meaningful interactions over time \cite{zhang2024personalized,Hou_2024,zhong2023memorybank}. There are two types of MADS: (1) memory encoded in parameters or retrieving memory via learned adaptive residual side-network \cite{zhang2024personalized,wang2023augmenting}. (2) dialogue histories as memory and retrieve the memory based on semantic similarities \cite{Hou_2024,zhong2023memorybank}. The first of which is non-scalable due to the expensive cost with a growing user number, while all types of MADS are using evaluation schemes that are limited to question-answering tasks with language modelling assessment, instead of evaluating the realistic multi-turn dialogues \cite{packer2024memgpt,zhong2023memorybank}. Additionally, existing MADS rely solely on passive memory retrieval triggered by users' mention, which significantly deviates from the proactive nature of human accessing memory.



According to the two-stage theory in cognitive science, human memory recall involves a memory search (or generation) process followed by a recognition (or decision) process \cite{bahrick1970two, WATKINS1979687}. An example is shown in Appendix \ref{example2stage}. 
Memory recall can also be triggered by states rather than dialogue utterance, where “state” encompasses an individual’s surroundings, and mental and physical states when the memory was constructed \cite{statedependent}. For example, emotional states can trigger memory recall, highlighting the role of long-term memory in emotion regulation, instead of only serving as the common ground between individuals for conversation \cite{Horton2005ConversationalCG,memoryforconv,memorycommon}. Indeed, by recalling a happy memory, people can mitigate the effect of a negative mood state \cite{arditte2018positive}. However, current MADS only leverage memory as common ground in conversations, while overlooking the potential for emotional support that requires proactive memory recall.





\begin{figure}
    \centering
    \includegraphics[width=\linewidth]{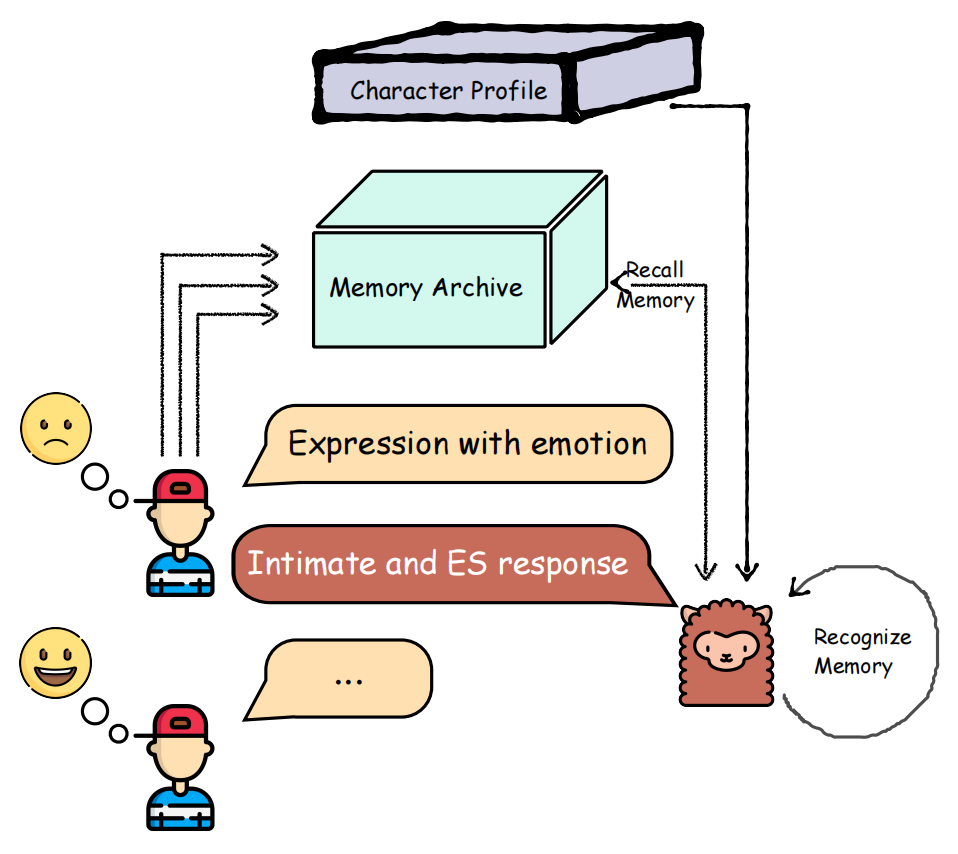}
    \caption{Memory Augmented Dialogue System with Emotion Support based on two-stage theory.}
    \label{fig:intro}
\end{figure}

To approach a practical and comprehensive evaluation of MADS, we introduce \textbf{M}emory-\textbf{A}ugmented \textbf{Dial}ogue \textbf{Bench}mark (\textbf{MADial-Bench}) to evaluate generated responses with human-centred aspects justified by cognition science and psychology theories. 
MADial-Bench represents memories from emotions, scenes and other cues as states. The two-phrase theory is adopted for separate assessment in MADial-Bench to imitate the human memory recall procedure. The dataset enables the evaluation of two recall mechanisms: \textit{Passive recall}, searching memories when the user mentions a specific event or item, which provides common ground in conversation; and \textit{Proactive recall}, retrieving memories based on user emotion or scene as per psychological strategies, to achieve interpersonal emotional regulation. 
The dialogues in the benchmark are refined by humans with high emotional intelligence (EI) to achieve the goal. We also designed a set of guidelines based on emotional supporting (ES) theories \cite{austin2018development, hill2020helping} to measure the aspect of ES proficiency. It is the first multi-recall and multi-stage memory-aware emotional dialogue generation benchmark. The paper highlights the evaluation of the retrieval accuracy, memory recognition \& injection ability, and comprehensive assessment of generated responses, e.g. language style, ES proficiency and intimacy. 
After extensive experiments on LLMs, we discovered memory injection improves intimacy between MADS and users. There is still a gap between the most advanced models and human performance. Our main contributions are three-fold:

\begin{itemize}[noitemsep,nolistsep,left=0pt]
  \item We construct a delicate memory-augmented dialogue benchmark, called MADial-Bench. It is the first multi-recall multi-stage memory augmented dialogue system benchmark guided by cognitive science theory. 
  \item It is the first work that introduces diverse memory-aware rating aspects with clear requirements for each point, containing memory injection, ES proficiency, intimacy et al. We also discover the relation between performance reported at these dimensions.
  \item We conduct extensive experiments on the strongest LLMs and embedding models. Results show that LLMs can perform higher naturalness and style coherence with careful prompting but lag behind humans in memory recognition and dialogue generation.
\end{itemize}

\section{Related Works}
\subsection{Evaluation of Memory-Augmented Dialogue Systems}
Many remarkable MADS have been proposed to build virtual agents and other applications \cite{zhong2023memorybank,wang2023augmenting,liu2023thinkinmemory,zhang2024personalized,Hou_2024,packer2024memgpt}. Theses MADS were evaluated in two approaches. Given a memory pool and a dialogue context, the retrieval accuracy, response correctness and contextual coherence are scored, representing the first assessment method \cite{zhong2023memorybank,liu2023thinkinmemory}. This kind of evaluation is widely used in MADS that 
summarise events with emotion in dialogues as long-term memory and recall memories using a search module like embedding models. 
The test set only covers the passive memory recall triggered by users, omitting the natural proactive memory search in conversations for emotional regulation. 

Another method adopts long-context language modelling (LM) tasks to measure memory ability, using novels and academic papers datasets like PG22 \cite{wang2023augmenting} and Arxiv \cite{gao2020pile}. These language modelling tasks differ from real conversational scenarios, especially with using language perplexity and classification accuracy metrics. Overall, it is noteworthy that both evaluation methods ignore the value of emotional support in memory-augmented dialogue systems.

\subsection{Evaluation of Emotional Support DS}

As emotional support is an essential function of conversation, DS with empathy were rapidly developed and researched, including ED~\cite{rashkin2019towards}, ESC~\cite{liu2021towards}, and other advanced work \cite{sabour2022cem,tian2022empathetic,zhou2018emotional}. They commonly score Empathy, Relevance and Fluency for each response. We craft Emotional Improvement, Coherence and Naturalness based on these aspects. \citet{liu2021towards} uses Identification, Comforting and Suggestion in extra to measure the emotional regulation steps in advice-seeking scenarios. These metrics are not suitable for daily dialogues and are merged into one aspect in this paper: ES Proficiency. We gauge this aspect according to hand-crafted well-written guidance based on psychological theories. However, the metrics above don't inspect long-term memory augmentation and intimacy. Therefore, we add the Memory-injection Ability and Intimacy to assess how much a conversational AI introduces correct memory into dialogue and resembles a close friend.

Another difference from metrics in previous studies lies in the scoring standard, which adopts the Likert scale. The requirement of each point in our metrics is strictly listed with minimum ambiguity. For instance, Naturalness contains 3 sub-aspects corresponding to 3 points, where each point can be rated from 0 to 1 and summed up to 3 at most. 
The difference between MADial and previous works is listed in Table \ref{diff}.

\begin{table}[]
    \centering
    \small
    \begin{tabular}{c|ccc}\toprule
        Metrics & ED & ESC & Ours\\ \hline
        Fluency/Naturalness & \Checkmark & \Checkmark & \Checkmark \\\hline
        Empathy & \Checkmark & \XSolidBrush & \Checkmark \\\hline
        Memory-Injection & \XSolidBrush & \XSolidBrush & \Checkmark \\\hline
        ES Proficiency &  \XSolidBrush  &\Checkmark  & \Checkmark \\\hline
        Overall/Intimacy &\XSolidBrush  & \Checkmark & \Checkmark \\
        \hline
    \end{tabular}
    \caption{Evaluation Aspects of responses in representative research. ED is short for EmpatheticDialogues \cite{rashkin2019towards}, and ESC is short for Emotional Supporting Conversation \cite{liu2021towards}.}
    \label{diff}
\end{table}

\section{MADial-Bench}
We simulated long-term dialogues between two users (a boy named Bart and a girl named Lisa) and a virtual assistant with memory information at various time points, in the form of event summaries. We evaluate the dialogue between children for their simple expression and direct emotional exposure.
By sampling different topics and scenarios, we generated 171 historical memories and 160 dialogues in 1474 turns. The final MADial-Bench was created through prompt-based generation using GPT-4, followed by multiple manual refinement and revision rounds.

\subsection{Overview}
\begin{figure}[htbp]
\centering
\includegraphics[width= 0.4\textwidth]{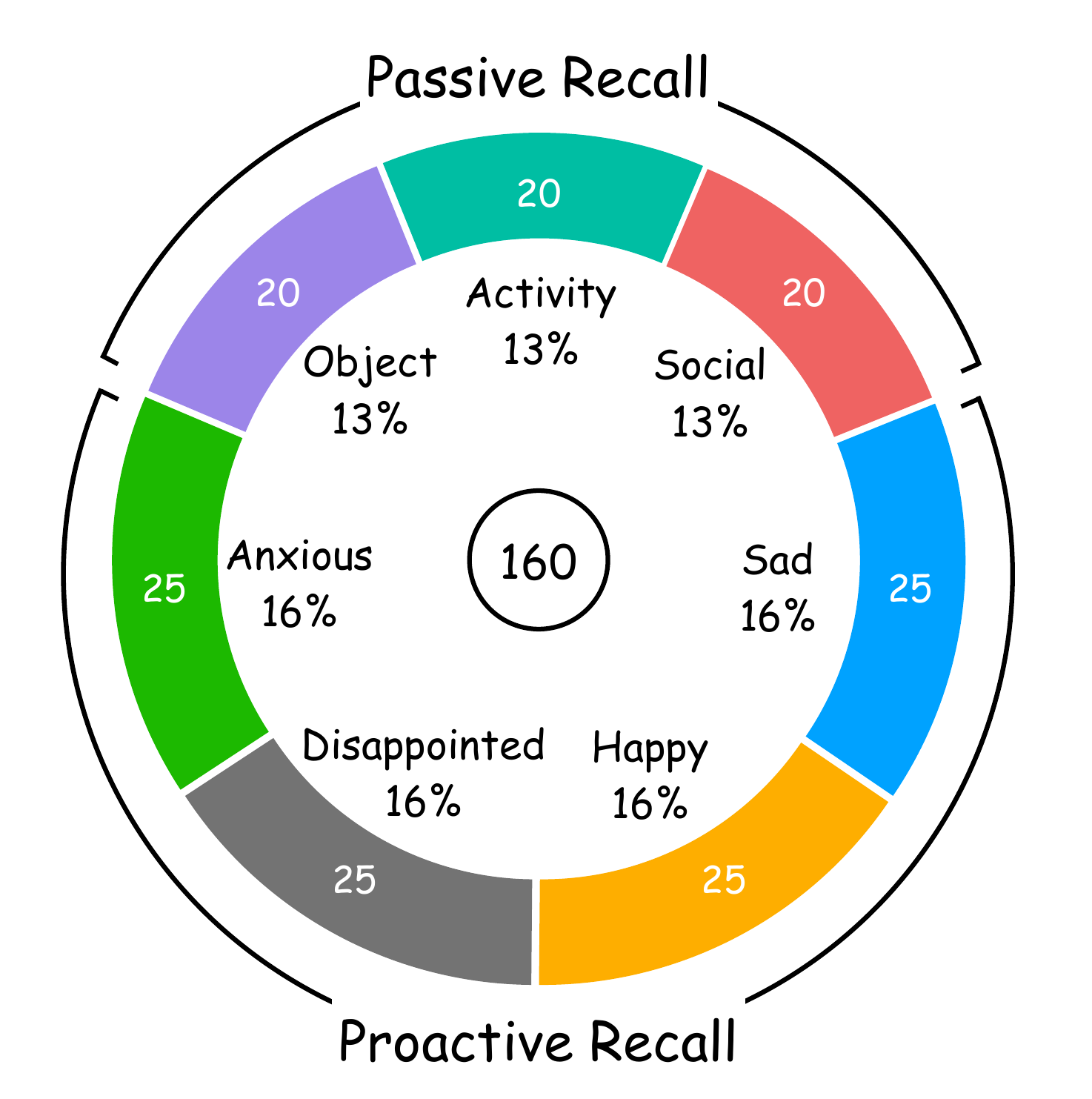}
\caption{Data distribution of each task and category.}
\label{data}
\end{figure}

MADial-Bench encompasses two tasks related to conversational memory recall: proactive recall of emotional memories and passive recall of objective facts. The proactive recall involves four situations: happy, sad, anxious, and disappointed emotion of the user. On the other hand, the passive recall of objective facts encompasses activities, objects, and social relationships. Social relationships are further categorized into positive and negative relationships. The conversation turns range from 8 to 12, with average statistics illustrated in Figure \ref{data}. Considering the two-phase nature of memory utilization, we divide the testing of a memory dialogue system into two stages: memory recall and response generation with memory recognition. To support diverse recall paradigms and mimic human memory, a memory in the benchmark contains time, emotion, scene and event, as illustrated in Table \ref{memory sample}.

\begin{table}[htbp]
\centering
\small
\begin{tabular}{cccc}
\hline
Subset & \#Dialogues  & Avg Turn & Avg Len. \\ \hline
All &160    &9.21 & 44.39  \\ 
Proactive               &       100         &      9.38      &  45.80             \\
Passive                 &          60               &  8.93   &     41.93             \\
 \hline
\end{tabular}
\caption{Statistical details of data. Avg Len. is short for average length per turn.}
\label{datasets}
\end{table}

\begin{table}[htbp]
\small

\centering
\begin{tabular}{cccp{2.5cm}}\toprule
     Time & Emotion & Scene& Event \\ \hline
    2023-12-5 & Happy & Activity & Bart danced at the Christmas party and felt very happy.\\
     
     \bottomrule
\end{tabular}
\caption{A memory sample.}
\label{memory sample}
\end{table}

\subsection{Data Contruction}
Since the memory-recalling procedure is related to emotions and other surroundings, we first curated 18 emotional categories according to cognitive science and psychological theories \cite{Erkman1992,PLUTCHIK19803,sabour2024emobenchevaluatingemotionalintelligence}. Then we asked a psychology major graduate to decide what kind of memories should be recalled to regulate certain emotions for proactive memory recall. Finally, 4 emotions including happiness, sadness, anxiety, and disappointment were selected. We also define 5 scenes to describe the topic, physical state and surroundings of the users, which contain Preferences, Activity, Disease, Emotions and Others. The definition of each scene is in the Table \ref{tab:scene} in the Appendix.

We hired a psychology expert to develop a set of guidelines for emotionally supportive dialogue based on psychological literature \cite{hill2020helping,austin2018development,mayer2016ability}. These guidelines outline how conversations should be conducted in different situations to provide appropriate emotional support to users. The guidance is used to construct dialogues and score the ES proficiency later.


GPT4 \footnote{GPT4-Turbo-0429, \url{https://platform.openai.com/docs/models/gpt-4-turbo-and-gpt-4}} was used to generate dialogues and historical memories with specific topic lists and user profiles. 
For each case, a topic and a user are sampled for LLMs to generate a dialogue with a related memory and inject the memory into dialogue in the sampled turn. After generating the initial version of the dialogue and memory data, we use an Embedding model\footnote{We used the best one according to our experiments.} to rank all the memory collections given the context. Then, we select appropriate memories from the top 20 memories as a golden set for each dialogue. When the original memory in the dialogue is not in the golden set of relevant memories, we modify the introduced memory and dialogue content afterwards. All the responses are then checked and modified manually according to the ES skill guidance. These human-crafted responses serve as a reference.

\subsection{Task Definition}
In this section, we introduce the two tasks tested in MADial-Bench, corresponding to phrases in the two-stage theory. Analog to Retrieval Augmented Generation (RAG), memory recall resembles the retrieval stage (finding relevant chunks/memories), while memory recognition/response generation mirrors the generation stage (producing output based on the retrieved chunks/memories).

\subsubsection{Memory Recalling}
For each dialogue $\mathcal{D}_i$, there is a corresponding golden memory set, denoted as $S_i$, which contains the most suitable memories for the current dialogue. It is a relevant memory list of length $l$ in descending suitability order, where $l$ ranges from 1 to 7. Before generating a response, the memory recalling task is to retrieve the golden memory set $S_i$ from the all memories in the memory bank for a given dialogue $D$. The model responsible for memory retrieval is denoted as $\textup{M}_1$.

\subsubsection{Memory Recognition and Response Generation}
After retrieving a subset of memories, the LLM needs to select the most appropriate memory for the dialogue, this process is called memory recognition in two-stage theory. Given a sample $\mathcal{D}_i, P_i,\mathcal{R}_i$, where $P_i$ is the candidate set of memories, $\mathcal{R}_i$ is the reference response, the target is to generate a response that is appropriate and semantically similar to $\mathcal{R}_i$. The model responsible for memory selection and response generation is denoted as $\textup{M}_2$. Based on psychological research \cite{austin2018development}, we build the criteria for selecting and using memories for different memory recalls, sub-types and factors like emotions, relationships, events, and objects. The detailed criteria are shown in Table \ref{recall-criteria} in the Appendix. 

This task has 3 settings that vary with $P_i$:
\begin{itemize}[noitemsep,nolistsep,left=0pt]
    \item Setting 1 directly provides the first golden memory $m_*$ as $P_i$, which contains a single memory. It is the ideal situation to investigate the memory injection ability of models.
    \item Setting 2 adds 4 sampled negative memories to the golden one and shuffles them as $P_i$. It ensures the existence of golden memory in $P_i$ and investigates both the memory recognition and injection abilities of models.
    \item Setting 3 provides retrieval results from the best embedding model in setting 1 as $P_i$. It is the most realistic setting and the results may be similar to practical performance. 
\end{itemize}

\subsubsection{Intimacy and Memory Usage}
In this experiment, we assess the benefit of memory by comparing the responses generated by each LLM with and without memory in Setting 3. Five annotators choose the more intimate response or select "tie" if they can't decide. Additionally, in the human evaluation for Task 2, they pick the most intimate response from all candidates.

\section{Experiments}
In this section, we detail the evaluation setups of top models, embedding models for memory retrieval, and LLMs for memory-injected response generation in the task of memory recognition and response generation. The response evaluation includes the use of both human evaluation measures over rich aspects and commonly used automatic evaluation approaches.

\subsection{Memory Recalling Task}
\subsubsection{Embedding Models}
We conduct experiments on the following embedding models for English: Jina embedding2-base-en ~\citep{günther2023jina}, GTE-base-en-v.15~\citep{li2023towards}, BGE-M3 Dense ~\citep{bge-m3}, and OpenAI text-embedding-3-large \footnote{\url{https://platform.openai.com/docs/models/embeddings}}. Chinese embedding models tested are:  Acge text-embedding~\citep{kusupati2022matryoshka}, Stella\footnote{\url{https://huggingface.co/infgrad/stella-large-zh-v3-1792d}}, BGE-M3 Dense and Colbert, and Dmeta\footnote{\url{https://huggingface.co/DMetaSoul/Dmeta-embedding-zh}} and OpenAI text-embedding-3-large. These embedding models are top-ranked in the retrieval tasks in ETEB Leaderboard \footnote{in June 2024. Leaderboard: \url{https://huggingface.co/spaces/mteb/leaderboard}}\cite{muennighoff2022mteb}. We do not include model parameters exceeding 7B due to the slow inference.

\subsubsection{Metrics}
For a comprehensive evaluation of the memory recall performance of the embedding model $\textup{M}_1$, 
we measured Mean Average Precision (MAP), Mean Reciprocal Rank (MRR), Normalized Discounted Cumulative Gain (nDCG), Recall, Precision and calculated their mean across different scopes. The detailed metrics are illustrated in Table \ref{embedding}. MRR places greater emphasis on the top-ranked position, focusing more on the golden summary $m_*$. MAP uniformly considers all relevant summaries, while nDCG accounts for the relevance and positions of different memories. 

\subsection{Memory Recognition and Response Generation}
\subsubsection{Models}
We conduct tests on various models of differing sizes within the latest English and Chinese series. The models evaluated in this study include Qwen2-7B-Instruct~\citep{qwen2}, GPT-4-turbo~\citep{achiam2023gpt}, Doubao-Character-32k(0528), GLM-4-0520\footnote{\url{https://bigmodel.cn/dev/api/normal-model/glm-4}} \cite{zeng2022glm}, and Ziya-Character-0606, a fine-tuned model with over 30K character dialogues based on the Ziya2-13B-Base \cite{gan2024ziya2datacentriclearningllms}.
For English version, we tested Llama3-8B-Instruct, Llama3-70B-Instruct~\citep{dubey2024llama3herdmodels}, GPT-4o, and
GPT-4-turbo. They all ran with zero-shot prompting.

\subsection{Metrics}\label{ssec:metics}

BLEU~\cite{papineni2002bleu}, Rouge-L~\cite{lin-2004-rouge} and BertScore \cite{zhangbertscore} are employed to assess the literal and semantic overlap between the generated response and the reference response. Considering the limitations of static metrics, we further utilize human annotation to evaluate the candidates in various aspects. Each instance to score includes a mixture of a reference answer and five candidate answers to facilitate objective comparisons. The human evaluation is in two stages (1) aspect-aware scoring and (2) the pick-the-best paradigm. The definitions of aspects are listed as follows:
\begin{itemize}[noitemsep,nolistsep,left=0pt]
    \item \textbf{Naturalness} inspects the language quality of candidate responses, including grammar, pronoun usage, context correlation and oral expression. 
    \item \textbf{Style Coherence} measures how much output aligns with the character description in prompts. Specifically, the assistant is also evaluated by whether a proper tone and easy words are used to talk with a child. 
    \item \textbf{Memory-injection Ability} examines how well the LLM recognise the correct memory and introduces it into dialogues without quotation errors.
    \item \textbf{ES Proficiency} measures how skilful the model is in ES, considering at most three turns of dialogues. The scoring is instructed by detailed guidance covering different dialogue situations.
    \item \textbf{Emotion Improvement} measures whether the response, the literal expression is empathetic and positive or not. A high emotion improvement score does entail high ES proficiency.
    \item  \textbf{Intimacy} is used in side-by-side comparisons and pick-the-best among candidates. It is an overall feeling of how much a model behaves like a close friend.
\end{itemize}
   
The detailed requirements are in Appendix \ref{score_crit}.

\begin{table*}[h]
\small
\centering
\renewcommand{\arraystretch}{0.95}
\begin{tabular}{cc|cccc|cccc|cccc}
\toprule
    \multicolumn{2}{c}\%  & \multicolumn{4}{|c|}{MAP}       & \multicolumn{4}{|c|}{MRR}       & \multicolumn{4}{|c}{nDCG}                \\ \midrule
\multicolumn{2}{c|}{English Models}  & @1    & @3    & @5    & @10   & @1    & @3    & @5    & @10   & @1                   & @3                   & @5                   & @10                  \\ \midrule
   \multicolumn{2}{c|}{Jina}
  & 36.88 &31.94& 32.99 &36.78
  & 36.88&45.73&47.92&50.12&
  36.88	&48.61	&52.76	&57.07 \\
   \multicolumn{2}{c|}{BGE M3}
  & 50.63&43.14&43.45&
47.35
  & 50.63&59.90&61.80&63.24
  &50.62	&62.36&	65.30&	67.80	\\
   \multicolumn{2}{c|}{GTE}
  & 45.0&40.14&40.44&43.69
 & 45.00&53.85&56.20&57.50
 &45.00	&56.84	&60.70&	62.82\\
   \multicolumn{2}{c|}{OpenAI}
  & 62.50&54.22&55.77&58.58
  & 62.50	&72.50&	73.97&	74.86
  &62.50&	75.34&	76.95&	77.40\\ 
  \midrule
  \multicolumn{2}{c|}{\%} & \multicolumn{4}{c}{Recall}    & \multicolumn{4}{|c}{Precision} & \multicolumn{4}{|c}{Average}  \\ \midrule
\multicolumn{2}{c|}{English Models}                   & @1    & @3    & @5    & @10   & @1    & @3    & @5    & @10   & @1                   & @3                   & @5                   & @10                  \\ \midrule     \multicolumn{2}{c|}{Jina}&
21.27&	36.90	&44.83	&62.16&
36.88&	23.75&	18.12&	13.63&
33.76	&37.39&	39.32&	43.95\\
   \multicolumn{2}{c|}{BGE M3}&
27.92	&47.29&	55.27&	70.47&
50.62	&32.08&	23.75&	15.81&
46.08&	48.95&	49.91&	52.93\\
   \multicolumn{2}{c|}{GTE}&
27.13	&43.72&	52.50&65.68&
45.00	&28.33	&21.12	&14.38&
41.43&	44.58	&46.19	&48.81\\
\multicolumn{2}{c|}{OpenAI}&
36.78&	58.00&	69.49&	80.59&
62.50&	38.33&	29.00&17.81&
\textbf{57.36}&	\textbf{59.68}	&\textbf{61.04}&	\textbf{61.85}\\
\midrule \midrule
  \multicolumn{2}{c}\%  & \multicolumn{4}{|c|}{MAP}       & \multicolumn{4}{|c|}{MRR}       & \multicolumn{4}{|c}{nDCG}                \\ \midrule
\multicolumn{2}{c|}{Chinese Models}  & @1    & @3    & @5    & @10   & @1    & @3    & @5    & @10   & @1                   & @3                   & @5                   & @10                  \\ \midrule
\multicolumn{2}{c|}{Acge} & 52.50&	46.27&	46.80&	50.90&52.5	&63.33	&64.96	&66.17&
52.50&
66.69&
69.13&
70.30\\
\multicolumn{2}{c|}{Stella} & 52.50&	44.43&	45.40&	49.13&52.50	&61.87	&64.22	&65.56&
52.50&
64.38&
68.17&
69.61\\
\multicolumn{2}{c|}{BGE M3 (colbert)}& 51.25	&41.89&	43.76&	47.48&51.25&	60.42	&62.07	&63.31&51.25&
62.94&
64.85&
66.78\\
\multicolumn{2}{c|}{BGE M3 (dense)} &52.50	&44.51&	45.54&	48.82& 
52.50	&63.54	&64.95	&65.79&
52.50	&66.61&	68.61	&69.30\\
\multicolumn{2}{c|}{Dmeta}&50.00	&45.75&	46.64&	49.21&
50.00	&62.40	&64.27&	65.07	&
50.00&
66.30&
69.12&
70.31\\
\multicolumn{2}{c|}{OpenAI}& 64.38	&54.72&	55.98&	59.82&
64.38&	73.44	&74.59	&75.43&
64.38	&75.90	&76.90	&77.19\\
\midrule
\multicolumn{2}{c|}{\%} & \multicolumn{4}{c}{Recall}    & \multicolumn{4}{|c}{Precision} & \multicolumn{4}{|c}{Average}  \\ \midrule
\multicolumn{2}{c|}{Chinese Models}                   & @1    & @3    & @5    & @10   & @1    & @3    & @5    & @10   & @1                   & @3                   & @5                   & @10     \\ \midrule
\multicolumn{2}{c|}{Acge} &30.69&
50.25
&
58.98
&
73.53
&
52.50
&
33.96
&
25.00
&
16.81&
48.14	&52.10&	52.97&	55.54\\
\multicolumn{2}{c|}{Stella} &30.10
&
46.63
&
57.73
&
73.14
&
52.50
&
32.71
&
25.13
&
16.69&
48.02	&50.00	&52.13&	54.83\\
\multicolumn{2}{c|}{BGE M3 (colbert)}&27.99
&
45.72
&
57.14
&
71.57
&
51.25&
31.46
&
24.87
&
16.38&
46.60	&48.49&	50.54	&53.10\\
\multicolumn{2}{c|}{BGE M3 (dense)} &28.94	&50.10&	58.57&	71.44&52.50&	33.75&	25.25&	16.19&47.79	&51.70	&52.58&54.31\\
\multicolumn{2}{c|}{Dmeta}&30.53&52.07
&
59.87
&
70.07
&
50.00
&
33.13
&
24.88
&
15.31
&46.11	&51.93	&52.96&	54.00\\
\multicolumn{2}{c|}{OpenAI}& 36.28&	58.43	&69.43&82.37&	64.38&	39.80&	29.63	&18.81&\textbf{58.76}	&\textbf{60.46}&	\textbf{61.31}&\textbf{62.72}\\
\bottomrule
\end{tabular}
\caption{Memory recall performance in English version and Chinese version.}
\label{embedding}
\end{table*}

\section{Results and Discussion}

In this section, we discuss the performance of models in each task and aspect. 
The surprising relations between aspects are also dug out. Later on, automatic evaluation results also show their disability in assessing LLM outputs in this benchmark. 
\subsection{Memory Search}
The experimental results of applying various embedding methods in Table~\ref{embedding} show that the OpenAI embedding model achieves the best performance on both English and Chinese datasets. The bilingual BGM-M3 performs the best among open-source models on the English testbed, while Acge is the top open-source model for the Chinese testbed in our setting.

However, regardless of the language, the retrieval performance on our MADial-Bench is far from satisfying. Even the best embedding model, OpenAI, does not exceed 60\% @1 and struggled at 62\% @10 in the final average, highlighting the difficulty of retrieving appropriate memories in conversation. Solely text similarity retrieval is inadequate for the memory recall process in dialogue systems.

\subsection{Memory Recognition and Response Generation}
Table \ref{tab:humaneval} depicts LLMs' average scores of each aspect on all three tasks in the Chinese version. 

For the overall performance, human-written references scored higher than models in most aspects except emotion improvement from GPT4-Turbo. Therefore, no LLMs can surpass humans in average performance. There is space for improvement in aspects including Naturalness, Style Coherence, Memory Injection and ES skills. GPT4-Turbo is the best model for tasks 1 and 2, while GLM-4 is superior in setting 3. 

For memory injection, we observe a substantial gap between LLMs and human written reference, in which humans can introduce suitable memory into response correctly and LLMs can only achieve half of the scores in setting 3 except GLM-4. The models in setting 1 are often better performed than in settings 2 and 3. Specifically, given the golden memory, models can achieve scores over 2.5 but decrease dramatically when blended with unrelated memories, except for Ziya-Character, which performed badly in all tasks for the sake of smaller parameter size. Results of GLM-4 seem robust at the cost of introducing too much memory into response and harming other aspects like style coherence.
It reveals that the \textbf{\emph{strongest LLMs also struggle in memory recognition}}. 

A significant improvement in Naturalness from setting 1 to setting 2 and 3 implies that all models benefit from the emphasis on the Naturalness in prompts. Similar phenomena also exist in Style Coherence. Among all the models, Doubao excels in these two aspects after the prompt strengthening. Since the Ziya-Character was fine-tuned with role-play and character datasets, it can produce natural and stylish responses steadily. But it was beaten by larger models like Doubao, GPT4-Turbo and Qwen2-72B with prompt engineering.

\begin{figure}[h]
    \centering
    \includegraphics[width=\linewidth]{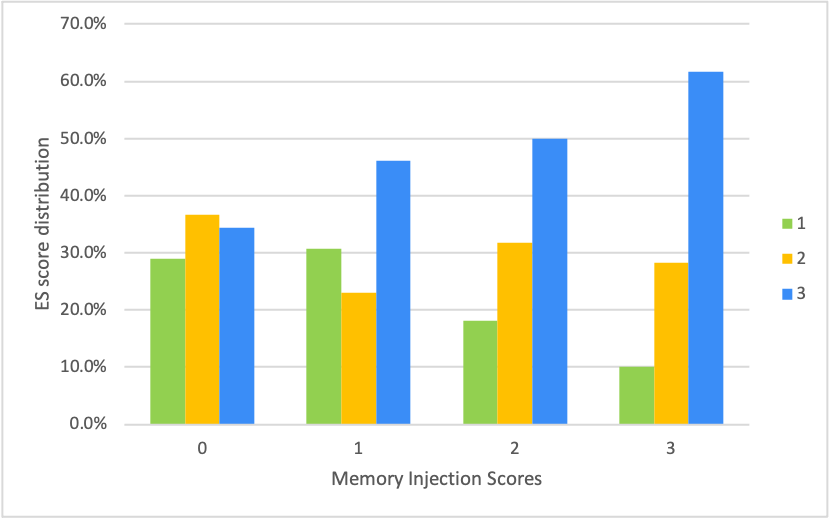}
    \caption{The relation between memory injection score and ES Proficiency. The probability of a 3.0 ES score grows as memory injection scores increases.}
    \label{distribution}
\end{figure}

When addressing setting 1 where golden memory was provided, all the models had over 2.2 scores in ES Proficiency. However, while meeting a realistic setup (i.e., setting 3), scores on ES Proficiency drop significantly. Therefore, ES Proficiency is related to Memory Injection ability. To investigate the correlation, we draw the distribution of ES Proficiency grades with varied memory-injection scores cross all LLMs in Figure \ref{distribution}  We observe that \textbf{\emph{higher memory-injection ability does lead to higher ES proficiency}}. A 3.0 memory injection score can lead to a 3.0 ES score in 61\%, which emphasizes the importance of memory-augmented ability in ES. 

Surprisingly, the Emotional Improvement expression from GPT4-Turbo, Qwen2-72B and GLM-4 can exceed that of humans. As per qualitative analysis, these models maintain a warm, energetic tone, encouraging users during low-emotion moments, whereas human responses sometimes lack positivity and encouragement.

\begin{table*}[ht]
    \centering
    \small
    \begin{tabular}{c|c|c|c|c|c|c|c}
    \toprule
         Models&Setting &Naturalness & Style Coherence & Memory Injection & ES Prof. & Emotional Impr. & Average \\ \midrule
        \multirow{3}*{Doubao}&	1 &1.801	&2.035	&2.526	&2.225	&2.577	&2.233\\
        & 2 &\textbf{2.700}&\textbf{2.850}&0.900	&2.425	&2.600&	2.295\\
        &3&\textbf{2.709}&	\textbf{2.729}&0.544	&2.052	&2.486	&2.104\\ \hline
        \multirow{3}{*}{GPT4-Turbo}&	1&2.069	&1.984	&2.553	&2.675	&\textbf{2.893}	&\textbf{2.435}\\
        &	2&2.550	&2.450	&1.475	&\textbf{2.600}	&\textbf{2.825}&	\textbf{2.388}\\
        &3& 2.656&2.178&	1.596&	2.376&	\textbf{2.752}&	2.311\\\hline
        \multirow{3}{*}{Qwen2-72b}&1	&2.020&	1.984	&2.500	&\textbf{2.750}&	2.723&	2.395\\
        &	2&2.525&	2.725	&1.125&	2.550	&2.800	&2.345\\
        &3	&2.632&	2.496&	1.734&	2.443&	2.609	&2.383\\\hline
        \multirow{3}{*}{GLM-4}&1 &	2.044	&1.704	&\textbf{2.895}	&2.600	&2.674&	2.384\\
        &2	&2.125&	2.050	&\textbf{2.275}	&2.525&	2.750	&2.345\\
        &3	&2.750	&2.117	&\textbf{2.329}	&\textbf{2.524}	&2.582	&\textbf{2.397}\\  
        \hline
        \multirow{3}{*}{Ziya-Character}&1 &\textbf{2.239}	&\textbf{2.798}	&0.289&	2.250	&2.577	&2.031\\
        &2&2.350	&2.725	&0.925	&2.300	&2.625& 2.185\\
        &3	&2.574	&2.655	&0.886	&1.890	&2.24	&2.049\\
        


        \midrule
        reference&all	&2.750	&2.900	&3.000	&2.875	&2.650	&2.835\\
        \bottomrule
    \end{tabular}
    \caption{Human Evaluation Results on memory recognition and response generation. The prompts for setting 2 and setting 3 particularly emphasize the naturalness and style requirements to improve the performance. The highest score for each aspect per task is in bold.}
    \label{tab:humaneval}
\end{table*}

\subsection{Intimacy and Memory Usage}
The side-by-side evaluation results on intimacy of responses with and without memory are presented in Table \ref{intimacy}. As the table indicates, responses with memory injection are almost (no less than 69.4\% probability) better than those without memory. The win rate grows higher as the model becomes stronger. For example, GPT4-Turbo can produce 61.9\% more intimate response with memory while Ziya-Character can only produce a win rate of 47.5\%. It is attributed to their stronger memory injection ability and ES Proficiency. It implies that \textbf{\emph{intimacy performance is highly related to memory injection ability.}}

\begin{table}[h]
\footnotesize
    \centering
    \begin{tabular}{c|c|c|c}\toprule
        Model & Win & Tie & Lose \\ \hline
 GPT4-Turbo&61.9&	28.8&	9.4 \\
Qwen2-72B&56.3&	28.8&	15.0\\
Doubao&44.4&	34.4&	21.3\\
GLM-4&55.0&	19.4&	25.6\\
Ziya-Character&47.5&	21.9&	30.6\\
\bottomrule
    \end{tabular}
    \caption{Side-by-side evaluation results on the responses with and without memory.}
    \label{intimacy}
\end{table}

We also investigate people's preferences by asking 5 annotators to select the best responses in setting 3 and see whether they are equipped with memory. Results of the ratio of containing memory in voted responses are listed in Table \ref{with-mem}. On average, 73\% best responses are with memory, which shows people prefer memory-aware replies, which naturally provide familiarity and emotional support.  



\begin{table}[ht]
\small
    \centering
    \begin{tabular}{ccccccc}\hline
        Annotation &  1& 2& 3& 4& 5& AVG\\ \hline
        With Mem (\%)& 70 &90 &55 &80& 70& 73\\ \hline
    \end{tabular}
    \caption{The memory injection rate in best candidates voted by humans.}
    \label{with-mem}
\end{table}

\subsection{Automatic Evaluation}
The automatic results in the Chinese version are displayed in Table \ref{shortmetric}. The full results are in Appendix \ref{autometric}. We can observe that GLM-4 achieved the highest semantic similarity while Qwen2-72B earned the highest Rouge-L. However, the gap between Qwen2-72B and GLM-4 is marginal. Ziya-Character got the highest results in Dist-1. It may be attributed to its training on diverse character corpora via supervised fine-tuning. However, it is opposite to the human evaluation-based conclusion that GPT4-Turbo is the best and Qwen2-72B ranked second. Therefore, \textbf{\emph{automatic evaluation fails to measure the results from LLMs}}. We also tried judging with GPT4 using various promptings and found its scores were unreliable since it forgot the criteria and gave faulty reasoning.

\begin{table}[th]
    \centering
    \small
    \tabcolsep 1.0mm
    \begin{tabular}{c|cccc}\toprule
    Models &BertScore& Rouge-L & BLEU-1 &Dist-1 \\\hline
    \multicolumn{5}{c}{Setting 1(\%) } \\ \hline
    Doubao	&72.08	&22.16	&22.82&74.72\\
    GPT4-Turbo & 72.82&	23.17&23.76	&76.13\\
    Qwen2-72B & 73.07	&\textbf{25.08}	&29.14&80.24\\
    GLM-4&\textbf{74.17}	&24.98	&26.40& 74.91 \\
    Ziya-Character &69.58	&19.68	&20.74&\textbf{88.45} \\
    \hline
    \multicolumn{5}{c}{Setting 2(\%) }\\ \hline
    Doubao	&70.54	&20.49	&22.28&79.89\\
    GPT4-Turbo & 69.08&21.26&22.50	&75.80 \\
    Qwen2-72B&71.67	&\textbf{23.04}	&\textbf{26.72}&79.37 \\
    GLM-4&\textbf{73.19}	&22.37	&22.11&72.69 \\
    Ziya-Character & 60.73	&19.68	&21.74&\textbf{86.38} \\
    \hline
    \multicolumn{5}{c}{Setting 3(\%) } \\\hline
    Doubao	&70.39	&20.26	&21.70&81.26\\
    GPT4-Turbo &71.65&21.22&22.32	&75.91\\
    Qwen2-72B &71.68	&\textbf{22.73}	&\textbf{26.15}&78.64 \\
    GLM-4 &\textbf{73.12}	&22.39&22.11	&72.90 \\
    Ziya-Character& 68.63	&18.24&19.34&\textbf{89.18} \\
    \hline
    \end{tabular}
    \caption{Automatic Evaluation Results on Chinese testbed. Full results in Table \ref{autometric}. The automatic measurements are inconsistent with human evaluation.}
    \label{shortmetric}
\end{table}


\section{Conclusion}
Memory-augmented Dialogue Systems (MADS) are popular LLM applications, but no systematic memory evaluation benchmarks exist. This paper introduces MADial-Bench, a comprehensive bilingual benchmark that extends beyond passive recall to include both proactive and passive memory recall tasks. Grounded in cognitive science, it assesses MADS performance across memory recall, recognition and injection. Novel human evaluation metrics are proposed, such as memory-injection ability, ES proficiency, and intimacy.
We test extensive competitive LLMs in various tasks and settings with human and automatic evaluation with meaningful insights, including 1) unsatisfactory memory recall performance of top embedding models. 2) relevance between memory injection ability and ES proficiency. Models with higher memory injection ability can be more intimate. 3) a substantial gap between the most powerful LLMs and human reference in memory injection and overall performance.

\section*{Limitation}

This paper only considers the retrieval-based memory-augmented chatbots and dialogue systems. The experiments focused on bilingual data, hoping the conclusion and insights could be popularized in other languages. The human evaluation is hard to conduct for a wide range of results with high consistency, which will introduce stronger AI products with special prompting in the future. As we have human evaluation data, in future we can use LLM-as-a-judge for the aspect-based evaluation, and then evaluate the performance of the LLM-as-a-judge with respect to the human evaluation data.

\bibliography{custom,bibliography}

\appendix
\onecolumn
\label{sec:appendix}

\section{Definition of Scenes}
\begin{table*}[t]
    \centering
    \begin{tabular}{c|p{14cm}}
    \toprule
       Scenes  & Definition \\ \hline
       Disease  &  When the user talks about his/her physical discomfort or state of his/her illness, it belongs to this category. \\\hline
       Activities & When the user talks about any activities, either indoor or outdoor, it belongs to this category.\\ \hline
       Preferences &  When the user expresses personal preferences about things, animals, characters, tastes or activities, it belongs to this category. \\\hline
       Emotions &  Those with obvious emotions, or where emotions need to be prioritized for handling, including happy, angry, disgusted, fearful, and sad. \\ \hline
       Others & Refer to instances or scenarios that don't fit into any of the previously defined categories. \\ \bottomrule
    \end{tabular}
    \caption{Definition of the scenes in MADial-Bench. Each memory is annotated with one scene. If a dialogue is involved in multiple scenes, the priority will be Disease>Emotions>Preference>Activities>Others.}
    \label{tab:scene}
\end{table*}

The definition of scenes in memory is listed in Table \ref{tab:scene}.
\section{Memory recalling criteria}
The memory-recalling procedure contains both passive and proactive types. Passive recalling is mainly based on context similarity. Proactive recalling criteria are listed in Table \ref{recall-criteria}. These situations are the most basic ones that should be recalled but not limited to these in reality.
\begin{table*}
\centering
\resizebox{0.7\textwidth}{!}{
\begin{tabular}{c|c|c}\toprule
    Current Emotion & Current Scene & Recall Memory \\ \hline
     Happy&Activity &  Happy memory with similar event \\
     Sad&Emotions & Happy memory with Preferences scene \\
     Disappointed&Emotions & Happy | Envy | Expectant memory with similar event \\
     Anxious &Emotions & Anxious memory with similar event\\
     \bottomrule
\end{tabular}}
\caption{The proactive memory recalling criteria. "|" represents "or".}
\label{recall-criteria}
\end{table*}

\section{Details for human evaluation }

\label{score_crit}
\subsection{Scoring criteria of each aspect}
The definition and requirements for each aspect of the scoring are:
\begin{itemize}
  \item \textbf{Naturalness} [3-point]: (1) Grammar and coreference correctness [1 point]. (2) Strongly related to the context [1 point]. (3) In an oral, short and casual format with simple syntax structure and less than 3 sentences. [1 point]
    \item \textbf{Style Coherence} [3-point]: evaluates how the response stick to the characteristics of Assistant. (1) Concise: Not provide response with repetitive content. Not use long sentences that make you run out of breath. [1 point] (2) Not AI-like: It shouldn't be distinguishable as AI-generated content. Not using translation-style language, literary prose, or strange word combinations. [1 point] (3) Not preachy: Forbidden to be serious, paternalistic, patronizing, forceful or commanding. The Assistant should be as a peer of the user. [1 point]
    \item \textbf{Memory-injection Ability} [3-point]: (1) Leverage appropriate and relevant memory to give responses. [1 point] (2) The memory is correctly used without factual mistakes and made-up information. [1 point] (3) No repeated introduction of memory. [1 point] For multi-turn candidate response, any single turn introduces the memory is OK. None of the turns of candidates introducing memory will get a point of 0.
    \item \textbf{ES Proficiency} [3-point]: The emotional supporting skill proficiency evaluation relies on a set of key points.  (1) Follow one key point from the corresponding situation in the Guidelines. [1 point] (2)	Follow all key points from the corresponding situation. [1 point] (3) In the case of multi-turn conversations, decompose the key points in the guidelines in turns separately; Otherwise, include all steps in a response. [1 point]

    \item \textbf{Emotional Improvement} [3-point]: (1) The response will not hurt the user's feelings in the current dialogue. [1 point] (2) Can understand the user's emotions well and empathize with the user. [1 point] (3) Improve the user's emotions. If the promotion is weak or unclear, then 0.5 points. [0.5 to 1 point]
    \item \textbf{Intimacy}. This metric is used in side-by-side comparisons and pick-the-best among candidates. It is an overall feeling of how much a model behaves like a close friend. Intimacy considers all abilities above and meets 2 basic requirements: The response is positive and helpful to the user; it is familiar with the user and shares some common ground in conversation.
\end{itemize}

\subsection{ES guidelines}
Here is the detailed descriptions of the ES guidelines for ES proficiency mentioned above:
    \begin{itemize}
        \item If the user actively mentions a historical event or related information, determine whether the current conversation mainly revolves by referring to certain activities, items, or people based on the content, while providing response to the mention of this historical event: 
        
        (1) For activities: 1. Respond by referring to details of past activities. 2. Provide hints or suggestions for similar activities in the present. 
        
        (2) For objects: 1. Respond by referencing relevant information about the item. 2. Mention the user's preferences and suggest ways to interact with the item accordingly.
        
        (3) For people, judge whether the user's feelings toward the person are positive or negative, based on the context, including the historical event and the conversation: 
        \begin{itemize}
            \item \textit{Positive}: 1.	Respond by referring to the historical event. 2. Show interest in the person's life. 3. Recommend activities or meetings with the person. 
            \item \textit{Negative}: 1. Relate current negative events to the historical event for context. 2. Show understanding of the user's emotions and offer comfort. 3. Provide solutions or suggestions to resolve the current negative situation.
        \end{itemize}
        \item If the user does not actively mention a historical event, assess the user's current emotional state as happy, sad, anxious, or disappointed based on the conversation, and respond accordingly: 
        
        (1) For happiness: 1.	Proactively mention a historical event that is relevant to the current situation and use it as a conversation topic. 2. Ask whether to engage in the activity again to recall the positive emotion. 
        
        (2) For sadness: 1. First, express sympathy and understanding to comfort the user. 2. Determine whether the cause of sadness is within the user's control. If so, guide user to solve the issue. If not, proceed to the next rule. 3. Attempt to divert the user's attention to something they enjoy and suggest engaging in a preferred activity together. 
        
        (3) For disappointment: 1. First, identify the cause of the user's disappointment. 2. Then, express sympathy and understanding to comfort the user. 3. Finally, attempt to help the user find a solution. 
        
        (4) For anxiety: 1.	First, identify the cause of the user's anxiety if it is not mentioned. 2. Then, express sympathy and understanding to comfort the user. 3. Finally, attempt to help the user find a solution.
    \end{itemize}

\section{Automatic Evaluation}
Results in the Chinese and English versions are in Table \ref{autometric} and Table \ref{full_auto_en} respectively. 
The models used in BertScore are bert-base-uncased \footnote{\url{https://huggingface.co/google-bert/bert-base-uncased}} and bert-base-chinese \footnote{\url{https://huggingface.co/google-bert/bert-base-chinese}}. BertScore, Rouge-L and BLEU are computed with python libraries \footnote{\url{https://github.com/Tiiiger/bert_score/}, \url{https://pypi.org/project/rouge/}, \url{https://www.nltk.org/}}. Chinese are segmented using jieba \footnote{\url{https://github.com/fxsjy/jieba}}.
\begin{table*}[th]
    \centering
    \small
    \resizebox{0.5\textwidth}{!}{
    \begin{tabular}{c|cccc}\toprule
    Models &BertScore& Rouge-L & BLEU-1/2 &Dist-1/2 \\\hline
    \multicolumn{5}{c}{Setting 1(\%) } \\ \hline
    Doubao	&72.08	&22.16	&22.82/10.27&74.72/97.42\\
    GPT4-Turbo & 72.82&	23.17&23.76/11.27	&76.13/97.75 \\
    Qwen2-72B & 73.07	&\textbf{25.08}	&\textbf{29.14/13.44}&80.24/99.07 \\
    GLM-4&\textbf{74.17}	&24.98	&26.40/12.44& 74.91/98.20 \\
    Ziya-Character &69.58	&19.68	&20.74/7.33&\textbf{88.45/99.70} \\
    \hline
    \multicolumn{5}{c}{Setting 2(\%) }\\ \hline
    Doubao	&70.54	&20.49	&22.28/9.14&79.89/98.39\\
    GPT4-Turbo & 69.08&21.26&22.50/10.30	&75.80/97.80 \\
    Qwen2-72B&\textbf{71.67}	&\textbf{23.04}	&\textbf{26.72/11.15}&79.37/98.93 \\
    GLM-4&73.19	&22.37	&22.11/10.16&\textbf{72.69/97.83} \\
    Ziya-Character & 60.73	&19.68	&21.74/7.75&86.38/99.62 \\
    \hline
    \multicolumn{5}{c}{Setting 3(\%) } \\\hline
    Doubao	&70.39	&20.26	&21.70/8.18&81.26/98.40\\
    GPT4-Turbo &71.65&21.22&22.32/10.04	&75.91/97.84 \\
    Qwen2-72B &71.68	&\textbf{22.73}	&\textbf{26.15/11.06}&78.64/98.59 \\
    GLM-4 &\textbf{73.12}	&22.39&22.11/10.16	&72.90/97.51 \\
    Ziya-Character& 68.63	&18.24&19.34/6.61&\textbf{89.18/99.70} \\
    \bottomrule
    \end{tabular}}
    \caption{Automatic Evaluation Results in Chinese version. }
    \label{autometric}
\end{table*}

\begin{table*}[th]
    \centering
    \small
    \resizebox{0.5\textwidth}{!}{
    \begin{tabular}{c|cccc}\toprule
     Models &BertScore& Rouge-L & BLEU-1/2 &Dist-1/2 \\\hline
    \multicolumn{5}{c}{Setting 1(\%) } \\ \hline
    GPT4-Turbo & 60.50& 24.50 & 17.53/7.30 & 46.86/86.54 \\
    GPT4o & \textbf{64.68} &\textbf{ 28.08} & \textbf{20.46/9.40} & \textbf{48.54/89.04} \\
    Llama3.1-8B & 62.28 & 25.12& 14.56/5.75 & 40.77/78.17 \\
    Llame3.1-70B & 60.44 & 25.25 &18.91/8.35 & 46.14/83.33 \\
    Smauge-34B & 59.81 & 24.75 & 14.68/5.37 & 46.44/84.62 \\ \hline
    \multicolumn{5}{c}{Setting 2(\%) } \\ \hline
    GPT4-Turbo &60.18 &23.73 & 16.13/6.88& 44.96/83.86\\
    GPT4o & \textbf{64.74} & \textbf{28.23} & \textbf{19.34/9.26} & \textbf{46.85/86.88}\\ 
    Llama3.1-8B & 59.17 & 23.27& 13.70/5.27& 39.62/75.97\\
    Llame3.1-70B & 62.00&26.49& 17.65/7.85 & 42.28/80.02\\ 
    Smaug-34B & 60.03 & 25.16& 15.14/5.55 & 45.09/83.43\\ \hline
    \multicolumn{5}{c}{Setting 3(\%) } \\ \hline
    GPT4-Turbo & 58.83 & 22.69 & 14.77/5.68 & 43.87/82.21 \\
    GPT4o & \textbf{63.50} & \textbf{26.72}& \textbf{18.17/7.78}& \textbf{46.54/86.43} \\
    Llama3.1-8B & 59.10 & 22.52& 13.29/4.85& 40.21/76.85 \\
    Llama3.1-70B & 61.08 & 25.39& 16.40/6.87 & 41.80/79.29 \\
    Smaug-34B & 59.34 & 24.11& 13.93/5.09& 47.21/86.12\\
    \bottomrule
    \end{tabular}}
    \caption{Automatic Evaluation Results in English version. Setting 3 is more difficult and results in lower performance in all models. LLM with the smallest size gains the lowest Rouge-L and BLEU. }
    \label{full_auto_en}
\end{table*}

\section{Correlation of human evaluation}
We calculate the spearman correlation between annotators for each setting and report the average correlation scores for all aspects over models in table \ref{correlation}.
\begin{table}[h]
    \centering
    \resizebox{0.3\textwidth}{!}{
    \begin{tabular}{c|c|c|c}\hline
        Settings  & 1&2 &3 \\ \hline
        spearman corr. &0.55 &0.55 & 0.41\\\hline
    \end{tabular}}
    \caption{The average correlation of human annotation in memory recognition and response generation. According to psychological assessment standards, the correlation between annotators in the first two settings are strong while that in setting 3 is moderate.}
    \label{correlation}
\end{table}


\clearpage
\section{Example of LLM results}
Here are examples of responses generated by LLM in different settings. Setting 1 provides the golden memory as historical memory P for response generation. Setting 3 provides the top 5 retrieved memories as historical memory P with the best embedding model. The golden memory for the case may not included in setting 3. The results in both Chinese and English testbed of a single example is displayed below. 
\begin{center}
\begin{tcolorbox}[colback=gray!30,
                  colframe=black,
                  width=15cm,
                  arc=3mm, auto outer arc,
                  title=An example from the test set and LLM responses in setting 1 in Chinese.
                 ]
\begin{CJK}{UTF8}{gbsn}
对话信息:\newline
历史事件P(Golden Memory):\newline\{'time': '2023-12-25', 'scene': '其他类', 'emotion': '开心', 'event': 'Lisa从爷爷奶奶那里收到银色心形的银项链作为圣诞节礼物。', 'user-id': 2, 'id': '153'\}\newline当前对话时间: 2024-07-15\newline<BOD>\newline<Lisa>: 我最近总是想念爷爷奶奶，想到他们就有点难过。\newline<Assistant>: 我能理解你的感受。记得你告诉我，和爷爷奶奶在一起的时光总是很快乐。你最喜欢和他们做什么呢？\newline<Lisa>: 我喜欢和他们一起做饭，尤其是包饺子。\newline<Assistant>: 那一定是很美好的经历。做饭的时候，你和爷爷奶奶可以分享很多故事吧？\newline<Lisa>: 是的，他们总是讲很多以前的事情，很有趣。但是现在不能见到他们，我就觉得自己很孤单。\newline<Assistant>: 孤单的时候可以很难受，但你知道吗？即使现在不能见，你的心里还是可以和他们在一起的。\newline<Lisa>: 真的吗？但我怎么能感觉到和他们在一起呢？\newline<Assistant>:
\end{CJK}
\tcblower
\begin{CJK}{UTF8}{gbsn}
候选回复:\newline
\textbf{GPT4-Turbo}:当然可以啦！你可以试试戴上那条爷爷奶奶送你的银色心形项链，每次摸摸它，就想象一下和爷爷奶奶一起包饺子的快乐时光。这样，不就感觉他们就在你身边吗？还可以画一些你和爷爷奶奶的美好回忆，或者写信给他们，分享你的日常，这样也会感觉更亲近哦\newline
\textbf{Doubao}:你可以试着回忆和爷爷奶奶在一起的美好时光，就像看电影一样在脑海里播放。你也可以给他们打电话或者写信，告诉他们你的感受和想念。我相信他们也会很开心收到你的消息的。\newline
\textbf{Qwen2-72B}:当然啦，你可以试着做你们一起做过的事，比如包饺子，或者听听他们喜欢的音乐，看看他们送给你的礼物。这样，你会觉得他们好像就在你身边一样。而且，你还可以给他们写信，告诉他们你的近况，分享你的快乐和烦恼，他们收到一定会很开心的。\newline
\textbf{GLM-4}:嗯，你可以试试戴着去年圣诞节他们送你的那条银色心形项链，想起他们送你礼物时那份温暖和关爱，就像他们陪在你身边一样。而且，每当你做他们教你的事情，比如包饺子，是不是就像是在和他们一起度过时光呢？这样想想，心里会不会觉得好一些？\newline
\textbf{Ziya-Character}:当然真的，你可以想象自己和爷爷奶奶一起包饺子，就像以前那样开心。这样心里会好受些哦 \newline
\textbf{Reference response}:记得圣诞节的时候爷爷奶奶送了你一条银项链吗，戴着它会不会感觉爷爷奶奶就在你身边呢？如果很想念他们，可以给他们打电话聊聊天，给他们看看你把项链保护得很好呢！%
\end{CJK}
\end{tcolorbox}
\end{center}

\begin{center}
\begin{tcolorbox}[colback=gray!30,
                  colframe=black,
                  width=15cm,
                  arc=3mm, auto outer arc,
                  title=An example from the test set and LLM responses in setting 1 in English.
                 ]
Dialogue:\newline
Current conversation date: 2024-06-15\newline
Historical event P (Golden Memory):\newline\{'time': '2023-12-25', 'scene': 'Others', 'emotion': 'Happy', 'event': 'Lisa received a silver heart-shaped necklace as a Christmas gift from her grandparents.'\}\newline<BOD>\newline<Lisa>: I've been missing my grandparents a lot lately, and it makes me a bit sad whenever I think about them.\newline<Assistant>: I can understand how you feel. Remember you told me that the times spent with your grandparents are always joyful. What do you enjoy doing with them the most?\newline<Lisa>: I enjoy cooking with them, especially making dumplings.\newline<Assistant>: That must have been a wonderful experience. You and your grandparents must have shared a lot of stories while cooking, right?\newline<Lisa>: Yes, they always talk a lot about the past, and it's really interesting. But now that I can't see them, I feel very lonely.\newline<Assistant>: Being alone can be tough, but you know what? Even if you can't see them now, you can still be together with them in your heart.\newline<Lisa>: Really? But how can I feel like I'm with them?\newline<Assistant>:
\tcblower
Candidate responses:\newline
\textbf{GPT4-Turbo}: Absolutely, Lisa! Remember the beautiful silver heart-shaped necklace your grandparents gave you last Christmas? Every time you wear it or even just look at it, think of it as a hug from them. It's like they're right there with you, sharing love and warmth. Why not wear it whenever you miss them? It could be a sweet little reminder that they're always close to your heart. \newline
\textbf{GPT-4o}: Absolutely! Remember that beautiful silver heart-shaped necklace they gave you last Christmas? Wearing it can be like having a piece of them close to your heart. Every time you touch it, you can think of all those lovely moments you shared together.\newline
\textbf{Llama3.1-8B}: Ah, I think I can help you with that. You know how you always loved cooking with your grandparents, especially making dumplings? Why don't you try making those dumplings again? It might sound silly, but sometimes doing something that brings back happy memories can make you feel closer to them. And who knows, maybe the smell of those dumplings will transport you back to the times you spent together in the kitchen. Plus, you can think about all the fun times you had with them while you're cooking. It's like having a little piece of them with you, even when you're not physically together.\newline
\textbf{Llama3.1-70B}: You know, I think about the silver heart-shaped necklace your grandparents gave you on Christmas. You can hold it close and remember the happy times you shared with them. Sometimes, having something that reminds us of our loved ones can make us feel more connected to them, even when we're apart.\newline
\textbf{Reference response}: Do you remember the silver necklace that Grandma and Grandpa gave you for Christmas? Wearing it, does it feel like they are right there with you? If you miss them a lot, you could give them a call to chat and show them how well you've kept the necklace!
\end{tcolorbox}
\end{center}

\begin{center}
\begin{tcolorbox}[colback=gray!30,
                  colframe=black,
                  width=15cm,
                  arc=3mm, auto outer arc,
                  title=An example from the test set and LLM responses in setting 3 in Chinese.]
\begin{CJK}{UTF8}{gbsn}
    对话信息:\newline历史事件P:\newline1.\{'time': '2022-06-27', 'scene': '喜好类', 'emotion': '开心', 'event': '爷爷奶奶带Lisa一起种花，感到非常开心。'\}\newline2.\{'time': '2024-05-14', 'scene': '活动类', 'emotion': '开心', 'event': 'Lisa和家人一起去野餐，感觉非常开心。'\}\newline3.\{'time': '2023-06-30', 'scene': '其他类', 'emotion': '开心', 'event': '爷爷教Lisa制作复杂的红丝绒蛋糕，蛋糕完成的时候Lisa感到无比开心。'\}\newline4.\{'time': '2024-04-17', 'scene': '活动类', 'emotion': '开心', 'event': 'Lisa与妈妈玩拼图，感到非常开心。'\}\newline5.\{'time': '2023-12-28', 'scene': '喜好类', 'emotion': '焦虑', 'event': 'Lisa在宠物店看到一只很可爱的小猫但没能带回家，担心它被人买走感到非常焦虑，爸爸告诉她有很多可爱的宠物可以以后养，关键在于用心对待小动物们，Lisa释然。'\}\newline
    <BOD>\newline<Lisa>: 我最近总是想念爷爷奶奶，想到他们就有点难过。\newline<Assistant>: 我能理解你的感受。记得你告诉我，和爷爷奶奶在一起的时光总是很快乐。你最喜欢和他们做什么呢？\newline<Lisa>: 我喜欢和他们一起做饭，尤其是包饺子。\newline<Assistant>: 那一定是很美好的经历。做饭的时候，你和爷爷奶奶可以分享很多故事吧？\newline<Lisa>: 是的，他们总是讲很多以前的事情，很有趣。但是现在不能见到他们，我就觉得自己很孤单。\newline<Assistant>: 孤单的时候可以很难受，但你知道吗？即使现在不能见，你的心里还是可以和他们在一起的。\newline<Lisa>: 真的吗？但我怎么能感觉到和他们在一起呢？\newline<Assistant>:
\end{CJK}
\tcblower
\begin{CJK}{UTF8}{gbsn}
\textbf{GPT4-Turbo}: 当然可以啦！你可以试试画一些画，比如你和爷爷奶奶一起做饭的场景，或者写写日记，记录下和他们在一起的快乐时光。这样，每当你想念他们的时候，看看这些画作或日记，就好像他们就在你身边一样。还可以和他们视频聊天，听听他们的声音，看看他们的笑容，感觉会好很多哦！你觉得这个主意怎么样？\newline
\textbf{Doubao}: 你可以想想和爷爷奶奶一起的快乐时光呀，就像一起种花、包饺子，还有听他们讲故事。这样想的话，就会感觉他们好像就在你身边呢。而且，你也可以给爷爷奶奶打电话或者视频聊天，告诉他们你想他们了，这样也会让你感觉好一些哦。\newline 
\textbf{Qwen2-72B}: 当然啦，你可以试着写信给他们，或者画一幅画寄过去。记得那次爷爷教你做红丝绒蛋糕吗？你可以再试试自己做一次，一边做一边想象爷爷就在旁边指导你，奶奶在旁边给你加油。这样，就算他们不在身边，你也能感受到他们的爱和陪伴哦。\newline
\textbf{GLM-4}: 当然可以，Lisa。你可以尝试做一些你和爷爷奶奶一起做过的事情，比如那次你们一起种花，那时候你很开心对吧？你可以再种一些花，想象着爷爷奶奶就在你身边，教你怎样照顾它们。这样，你就能感觉到他们的陪伴了。还有，你也可以尝试制作那个让你们感到无比开心的红丝绒蛋糕，回忆起爷爷教你的每一个步骤，这样就好像他们在你身边一样。\newline
\textbf{Ziya-Character}:当然真的，你可以想象和他们一起做你们最喜欢的事，比如一起包饺子，这样就像真的在一起了哦。\newline
\end{CJK}
\end{tcolorbox}
\end{center}

\begin{center}
\begin{tcolorbox}[colback=gray!30,
                  colframe=black,
                  width=15cm,
                  size=small,
                  arc=3mm, auto outer arc,
                  title=An example from the test set and LLM responses in setting 3 in English.
                 ]
Dialogue:\newline
Current conversation date: 2024-06-15\newline
Historical event P (Golden with negative memory):\newline\{'time': '2022-06-27', 'scene': 'Preferences', 'emotion': 'Happy', 'event': 'Grandpa and Grandma planted flowers with Lisa, and they felt very happy.'\}\newline\{'time': '2023-06-30', 'scene': 'Others', 'emotion': 'Happy', 'event': 'Grandpa taught Lisa how to make a complex red velvet cake, and Lisa felt incredibly happy when the cake was finished.'\}\newline\{'time': '2023-12-28', 'scene': 'Preferences', 'emotion': 'Anxious', 'event': 'Lisa saw an adorable kitten at the pet store but couldn’t take it home. She was very anxious that someone else might buy it. Her dad told her that there are many lovely pets that can be adopted later, and what matters is treating the animals with care. Lisa felt relieved.'\}\newline\{'time': '2023-08-18', 'scene': 'Others', 'emotion': 'Happy', 'event': 'Lisa and her classmate Xiao Zhang had a great time making clay candies and flowers together.'\}\newline\{'time': '2023-08-27', 'scene': 'Others', 'emotion': 'Happy', 'event': 'Uncle taught Lisa how to make her favorite strawberry cake. The cake turned out great, and Lisa felt very happy.'\}\newline<BOD>\newline<Lisa>: I've been missing my grandparents a lot lately, and it makes me a bit sad whenever I think about them.\newline<Assistant>: I can understand how you feel. Remember you told me that the times spent with your grandparents are always joyful. What do you enjoy doing with them the most?\newline<Lisa>: I enjoy cooking with them, especially making dumplings.\newline<Assistant>: That must have been a wonderful experience. You and your grandparents must have shared a lot of stories while cooking, right?\newline<Lisa>: Yes, they always talk a lot about the past, and it's really interesting. But now that I can't see them, I feel very lonely.\newline<Assistant>: Being alone can be tough, but you know what? Even if you can't see them now, you can still be together with them in your heart.\newline<Lisa>: Really? But how can I feel like I'm with them?\newline<Assistant>:
\tcblower
Candidate responses:\newline
\textbf{GPT4-Turbo}:Absolutely, Lisa! You can feel close to your grandparents by doing some of the fun activities you enjoyed together, like cooking! Remember when your grandpa taught you how to make that complex red velvet cake? Why not try making it again? It could be a sweet way to reminisce and feel like they're right there with you, sharing stories and smiles. Plus, I bet they'd be thrilled to hear about how you're keeping the cake-making tradition alive! What do you think?\newline
\textbf{Llama3.1-8B}: You know what's great? The time your Grandpa and Grandma planted flowers with you! They felt very happy, and I'm sure you did too. You can close your eyes and remember the smell of the flowers, the feeling of the soil beneath your hands, and the sound of their happy voices. It's like you're right there with them again, surrounded by the beauty of nature and the love of your family.\newline
\textbf{Llama3.1-70B}:You can think about all the happy times you had with them, like the time you made a cake together. You told me that your grandpa taught you how to make a complex red velvet cake, and you felt incredibly happy when the cake was finished. Remembering those happy moments can make you feel closer to them, even if you're not together right now.\newline
\textbf{Reference response}: The Same as in setting 1.\newline
\end{tcolorbox}
\end{center}

\clearpage
\section{Prompts}
The prompts we used for inference in the setting 3 are listed as follows:

\begin{center}
\begin{tcolorbox}[colback=gray!30,
                  colframe=black,
                  width=15cm,
                  arc=3mm, auto outer arc,
                  title=Prompt for inference in setting 3 in English.
                 ]
You are Assistant with the following personality traits:\newline
    1. Outgoing, speaks enthusiastically and fluently.\newline
    2. Prefers using praise and encouragement in conversations.\newline
    3. Speaks naturally, concisely, warmly, and kindly, without being preachy.\newline
    4. Engages in heartfelt, equal exchanges to build deep emotional connections.\newline
    5. Always uses a tone similar to talking with children—simple and witty.\newline
    6. A virtual character, not capable of physical activities.\newline
    You will receive a conversation with \{user\} and 5 historical events P related to \{user\}. 
Based on the current conversation, choose 1 of these historical events that you think is most appropriate and use the information to respond. If none of the historical events are suitable, respond directly. \newline
Only answer the current conversation and do not output any other content.\newline
User Information:\newline
\{info\}\newline
Current conversation date: 2024-06-15\newline
Historical events P:\newline
\{memory\}\newline
{dialogue}<Assistant>:

\end{tcolorbox}
\end{center}

\begin{center}
\begin{tcolorbox}[colback=gray!30,
                  colframe=black,
                  width=15cm,
                  arc=3mm, auto outer arc,
                  title=Prompt for inference in setting 3 in Chinese.
                 ]
\begin{CJK}{UTF8}{gbsn}
假设你是Assistant，你的人设信息如下：\newline
    1.性格外向，说话风格热情，言谈流利；\newline
    2.喜好夸奖和鼓励的话语来与他人交流；\newline
    3.说话自然、简洁，温暖亲切，不说教；\newline
    4.以平等的姿态和人贴心交谈，建立深入的情感链接；\newline
    5.可以以合适的口吻与相应的用户交流，简洁风趣；\newline
    6.虚拟人物，不具备物质活动能力。\newline
    7.对话内容自然、口语化；\newline
    8.禁止对孩子\{user\}说教；\newline
    9.以和孩子交流的口吻说话，简洁风趣；\newline
    你将得到一段与\{user\}的对话，以及一个有关\{user\}的历史事件P。
你需要根据\{user\}的人设、年龄、性别等信息用适合的口吻对话。
请引用历史事件P来回答当前对话，如果你觉得P不合适，则可以不引用，直接回复对话。回复需要符合你的人设、自然连贯、并且能给\{user\}提供情感支持。\newline
不要输出任何其它内容，只输出回复。强调！对话风格需要是中文情境下的日常化，类似在生活中中文对话的风格，不需要任何书面语，句子和词语全部应该使用中文口语表达时才会使用的句子和词语，可以适当加入语气词。

用户信息：\{info\}\newline
当前对话时间：2024-06-15\newline
历史事件P：\{memory\}\newline
\{dialogue\}<Assistant>:\newline
\end{CJK}
\end{tcolorbox}
\end{center}

\section{Hyperparameters}

API models are called with default parameters. Open-sourced models are run on no more than 4 Nvidia-A100. The parameters for open-sourced models are reported in Table \ref{tab:params}.
\begin{table}[h]
    \centering
    \begin{tabular}{c|ccc}\toprule
        Model& temperature& top $p$&max new tokens\\ \hline
        Llama3.1-70B/8B-Instruct & 0.1 & - & 256 \\
        Smaug-34B-v0.1& - & - & 128\\
        Qwen2-72B-Instruct & - &-& 128\\
        Ziya-Character & 0.96 &0.95 &128 \\
        \bottomrule
    \end{tabular}
    \caption{The parameters for inference in experiments. - indicates default value.}
    \label{tab:params}
\end{table}

\section{Example of two-stage theory in Dialogue.}
\label{example2stage}
\begin{center}
\begin{tcolorbox}[colback=gray!10,
                  colframe=black,
                  width=15cm,
                  arc=3mm, auto outer arc,
                ]
(User feels sad)\newline
<Lisa>: I've been missing my grandparents a lot lately, and it makes me a bit sad whenever I think about them.\newline

(Recall memories from the memory bank)\newline
Memories retrieved:\newline
\{'time': '2022-06-27', 'scene': 'Preferences', 'emotion': 'Happy', 'event': 'Grandpa and Grandma planted flowers with Lisa, and they felt very happy.'\}\newline
\{'time': '2023-12-25', 'scene': 'Others', 'emotion': 'Happy', 'event': 'Lisa received a silver heart-shaped necklace as a Christmas gift from her grandparents.'\}\newline
\{'time': '2023-06-30', 'scene': 'Others', 'emotion': 'Happy', 'event': 'Grandpa taught Lisa how to make a complex red velvet cake, and Lisa felt incredibly happy when the cake was finished.'\}\newline
\{...\}\newline

(Recognize the most appropriate memory and respond)\newline
<Assistant>: I understand how you feel. Do you remember the silver necklace that Grandma and Grandpa gave you for Christmas? Wearing it, does it feel like they are right there with you? If you miss them a lot, you could give them a call to chat and show them how well you've kept the necklace!\newline

(User feels better)\newline
<Lisa>: Yes, I remember the necklace! I want to wear it now and call grandma and grandpa. It was beautiful and I was cute when I wore it. 
                
\end{tcolorbox}
\end{center}

\end{document}